%% file: mldotnet.tex
\documentclass[sigconf,screen]{acmart}

\usepackage{microtype}
\usepackage{graphicx}
\usepackage{subcaption}
\usepackage{booktabs} 

\usepackage{xspace}
\usepackage{listings} 
\usepackage{balance}
\usepackage{flushend}

\def\BibTeX{{\rm B\kern-.05em{\sc i\kern-.025em b}\kern-.08emT\kern-.1667em\lower.7ex\hbox{E}\kern-.125emX}}
    
\newcommand{\mi}[1]{[\textcolor{red}{MI: #1}]}
\newcommand{\mw}[1]{[\textcolor{blue}{MW: #1}]}

\newcommand{\eat}[1]{}
\newcommand{\at}[1]{\protect\ensuremath{\mathsf{#1}}\xspace}
\newcommand{\stitle}[1]{\vspace{0.0ex}\noindent{\bf #1}}
\newcommand{\etitle}[1]{\vspace{0.8ex}\noindent{\underline{\em #1}}}

\newcommand{\msft}{Microsoft\xspace}
\newcommand{\tool}{\textsc{ML.NET}\xspace}
\newcommand{\skl}{Scikit-learn\xspace}
\newcommand{\sk}{Sklearn\xspace}
\newcommand{\idataview}{\at{DataView}}
\newcommand{\idataviews}{\at{DataViews}}

\definecolor{dkgreen}{rgb}{0,0.6,0}
\definecolor{gray}{rgb}{0.5,0.5,0.5}
\definecolor{mauve}{rgb}{0.58,0,0.82}

\lstdefinelanguage{Scala}{
  keywords={typeof, new, true, false, catch,def,val, function, return, null, catch, switch, var, if, in, while, do, else, case, break},
  keywordstyle=\color{blue}\bfseries,
  ndkeywords={class, export,extends, boolean, throw, implements, import, this, abstract},
  ndkeywordstyle=\color{dkgreen}\bfseries,
  otherkeywords={+, =>,<=, ==, >,< , ||},
  identifierstyle=\color{black},
  sensitive=false,
  comment=[l]{//},
  morecomment=[s]{/*}{*/},
  commentstyle=\color{purple}\ttfamily,
  stringstyle=\color{red}\ttfamily,
  morestring=[b]',
  morestring=[b]",
  moredelim=**[is][\color{red}]{@}{@},
}
\lstdefinestyle{fault}{ numbers=none, xleftmargin=1.5em , otherkeywords={ =>,<=, ==, > , ||}}

\copyrightyear{2019}
\acmYear{2019}
\setcopyright{acmlicensed}
\acmConference[KDD '19]{The 25th ACM SIGKDD Conference on Knowledge
Discovery and Data Mining}{August 4--8, 2019}{Anchorage, AK, USA}
\acmBooktitle{The 25th ACM SIGKDD Conference on Knowledge Discovery and
Data Mining (KDD '19), August 4--8, 2019, Anchorage, AK, USA}
\acmPrice{15.00}
\acmDOI{10.1145/3292500.3330667}
\acmISBN{978-1-4503-6201-6/19/08}

\begin{document}

%
\title{Machine Learning at Microsoft with ML.NET}

%
\author{Zeeshan Ahmed$^1$, Saeed Amizadeh$^1$, Mikhail Bilenko$^2$, Rogan Carr$^1$, Wei-Sheng Chin$^1$, Yael Dekel$^1$, Xavier Dupre$^1$, Vadim Eksarevskiy$^1$, Eric Erhardt$^1$, Costin Eseanu$^1$*, Senja Filipi$^1$, Tom Finley$^1$, Abhishek Goswami$^1$, Monte Hoover$^1$, Scott Inglis$^1$, Matteo Interlandi$^1$,Shon Katzenberger$^1$, Najeeb Kazmi$^1$, Gleb Krivosheev$^1$, Pete Luferenko$^1$}\authornote{Pete Luferenko is now at Facebook. Costinn Eseanu and Gal Oshri are now at Google. Sarthak Shah is now at Pinterest.}
\author{Ivan Matantsev$^1$, Sergiy Matusevych$^1$, Shahab Moradi$^1$, Gani Nazirov$^1$, Justin Ormont$^1$, Gal Oshri$^1$*, Artidoro Pagnoni$^1$, Jignesh Parmar$^1$, Prabhat Roy$^1$, Sarthak Shah$^1$*, Mohammad Zeeshan Siddiqui$^1$, Markus Weimer$^1$, Shauheen Zahirazami$^1$, Yiwen Zhu$^1$}
\affiliation{$^1$Microsoft, $^2$Yandex}

%
\renewcommand{\shortauthors}{Ahmed, et al.}

%


\begin{abstract}
Machine Learning is transitioning from an art and science into a technology available to every developer.
In the near future, every application on every platform will incorporate  trained models to encode data-based decisions that would be impossible for developers to author. This presents a significant engineering challenge, since currently data science and modeling are largely decoupled from standard software development processes.
This separation makes incorporating  machine learning capabilities inside applications unnecessarily costly and difficult, and furthermore discourage developers from embracing ML in first place. 
%

In this paper we present \tool, a framework developed at \msft over the last decade in response to the challenge of making it easy to ship machine learning models in large software applications.  We present its architecture, and illuminate the application demands that shaped it. 
Specifically, we introduce \idataview, the core data abstraction of \tool which allows it to capture full predictive pipelines efficiently and consistently across training and inference lifecycles.
We close the paper with a surprisingly favorable performance study of \tool compared to more recent entrants, and a discussion of some lessons learned. 
\end{abstract}


%
\maketitle




\input{introduction}

\input{motivation}
\input{design}

\input{implementation}

\input{experiments}

\input{lessons}
\input{related}

\input{conclusions}


\bibliographystyle{ACM-Reference-Format}
\bibliography{mldotnet}

\newpage

\input{reproducibility}
\end{document}

%% file: introduction.tex
\section{Introduction}
\label{sec:introduction}
\vspace{-1ex}
We are witnessing an explosion of new frameworks for building Machine Learning (ML) models~\cite{tensorflow,cntk,pytorch,mxnet,scikit,michelangelo,trans,h2o}.
This profusion is motivated by the transition from machine learning as an art and science into a set of technologies readily available to every developer. 
An outcome of this transition is the abundance of applications that rely on trained models for functionalities that evade traditional programming due to their complex statistical nature. Speech recognition and image classification are only the most prominent such cases.
This unfolding future, where most applications make use of at least one model, profoundly differs from the current practice in which data science and software engineering are performed in separate and different processes and sometimes even organizations. Furthermore, in current practice, models are routinely deployed and managed in completely distinct ways from other software artifacts. While typical software packages are seamlessly compiled and run on a myriad of heterogeneous devices, machine learning models are often relegated to be run as web services in relatively inefficient containers~\cite{clipper,tf-serving2,DBLP:conf/osdi/LeeSCSWI18}. 
This pattern not only severely limits the kinds of applications one can build with machine learning capabilities~\cite{DBLP:journals/debu/LeeSCWI18}, but also discourages developers from embracing ML as a core component of applications.

At \msft, we have encountered this phenomenon across a wide spectrum of applications and devices, ranging from services and server software to mobile and desktop applications running on PCs, Servers, Data Centers, Phones, Game Consoles and IOT devices. A machine learning toolkit for such diverse use cases, frequently deeply embedded in applications, must satisfy additional constraints compared to the recent cohort of toolkits. For example, it has to limit library dependencies that are uncommon for applications; it must cope with datasets too large to fit in RAM; it has to scale to many or few cores and nodes; it has to be portable across many target platforms; it has to be model class agnostic, as different ML problems lend themselves to different model classes; and, most importantly, it has to capture the full prediction pipeline that takes a test example from a given domain (e.g., an email with headers and body) and produces a prediction that can often be structured and domain-specific (e.g., a collection of likely short responses).
The requirement to encapsulate predictive pipelines is of paramount importance because it allows for effectively decoupling application logic from model development.  Carrying the complete train-time pipeline into production provides a dependable way for building efficient, reproducible, production-ready models~\cite{google-rules-of-ml}.

The need for ML pipelines has been recognized previously.  Python libraries such as \skl~\cite{scikit} provide the ability to author complex machine learning cascades. Python has become the most popular language for data science thanks to its simplicity, interactive nature (e.g., notebooks~\cite{jupyter,zeppelin}) and breadth of libraries (e.g., numpy~\cite{numpy}, pandas~\cite{pandas}, matplotlib~\cite{matplot}).
However, Python-based libraries inherit many syntactic idiosyncrasies and language constraints (e.g., interpreted execution, dynamic typing, global interpreter locks that restrict parallelization), making them suboptimal for high-performance applications targeting a myriad of devices.

In this paper we introduce \tool: 
a machine learning framework allowing developers to author and deploy in their applications complex ML pipelines composed of data featurizers and state of the art machine learning models. 
Pipelines implemented and trained using \tool can be seamlessly surfaced for prediction without any modification: training and prediction, in fact, share the same code paths, and adding a model into an application is as easy as importing \tool runtime and binding the inputs/output data sources.
\tool's ability to capture full, end-to-end pipelines has been demonstrated by the fact that
1,000s of \msft's data scientists and developers have been using \tool over the past decade, infusing 100s of products and services with machine learning models used by hundreds of millions of users worldwide.

\tool supports large scale machine learning thanks to an internal design borrowing ideas from relational database management systems and embodied in its main abstraction: \idataview. \idataview provides \emph{compositional processing} of \emph{schematized data} while being able to \emph{gracefully and efficiently} handle \emph{high dimensional} data in datasets \emph{larger than main memory}. Like views in relational databases, a \idataview is the result of computations over one or more base tables or views, is immutable and lazily evaluated (unless forced to be materialized, e.g., when multiple passes over the data are requested).
Under the hood, \idataview provides 
streaming access to data so that working sets can exceed main memory.  
%
\eat{
\tool embraces a strongly and statically typed API and trades interactivity with usability: 
thanks to this design, most of the errors can be intercepted at compile time rather than at runtime. 
\tool comes with 
Python bindings mirroring \skl interface for data scientists that are familiar with the latter. 
}

We run an experimental evaluation comparing \tool with \skl and H2O~\cite{h2o}. To examine runtime, accuracy and scalability performance, we set up several experiments over three different datasets and utilizing different data sample rates. 
Our experiments show that \tool outperforms both \sk and H2O in speed and accuracy, in most cases by a large margin. 

Summarizing our contributions are:
\begin{itemize}

\item The introduction of \tool: a machine learning framework for authoring production-grade machine learning pipelines, which can then be easily integrated into applications running on heterogeneous devices;

\item Discussion on the motivations pushing \msft to develop \tool, and on the lessons we learned from helping thousands of developers in building and deploying ML pipelines at enterprise and cloud scale; 

\item Introduction of the \idataview abstraction, and how it translates into efficient executions through streaming data access, immutability and lazy evaluation;

\item A set of experiments comparing \tool against the well known \skl and the more recent H2O, and proving \tool's state-of-the-art performance.
\vspace{-1ex}
\end{itemize}

The remainder of the paper is organized as follows: Section~\ref{sec-motivation} introduces the main motivations behind the development of \tool. Sections~\ref{sec-design} introduces \tool's design and the \idataview abstraction, while Section~\ref{sec-implementation} drills into the details of \tool implementation. Section~\ref{sec-experiments} contains our experimental evaluation. Lessons learned are introduced in Section~\ref{sec-lessons}. The paper ends with related works and conclusions, respectively in Sections~\ref{sec-related} and~\ref{sec-conclusions}.

%% file: motivation.tex
\section{Motivations and Overview}
\label{sec-motivation}

The goal of the \tool project is improving the development life-cycle of ML pipelines.  While pipelines proceed from the initial experimentation stages to engineering for scaling up and eventual deployment into applications, they traditionally require significant (and sometimes complete) rewriting at significant cost. 
Aiming to reduce such costs by simplifying and unifying the underlying ML frameworks, we observed three interesting patterns:

\etitle{Pattern 1:} Many data scientists within \msft were not following the interactive pattern made popular by notebooks and Python ML libraries such as \sk. 
This was due to two key factors: the sheer size of production-grade datasets and accuracy of the final models.
While an ``accurate enough'' model can be easily developed by iteratively refining the ML pipeline on data samples, finding the best model often requires experimenting over the entire dataset. In this context, interactive exploration is of limited applicability inasmuch as most of the datasets were large enough to not fit into main memory on a single machine.
Working on large datasets has led to many of \msft's data scientists working in batch mode:  running multiple scripts concurrently sweeping over models and parameters, and not performing much exploratory data analysis. 

\etitle{Pattern 2:} In order to obtain the best possible model, data scientists focused their experiments on  several state of the art algorithms from different model families. These methods were typically developed by different groups, often coming directly from researchers within our company. 
As a result, each model was originally implemented without a standard API, input format, or hyperparameter notation. Data scientists were therefore spending considerable effort on implementing glue code and ad-hoc wrappers around different algorithms and data formats to employ them in their pipelines.
In general, getting all of the steps right required multiple iterations and significant time costs. Because ad-hoc pipelines are constructed to run in batch mode, there is no static, compile-time checking to detect any inconsistencies in dataflow across the pipelines.  As a result,  pipeline debugging was performed via log parsing and errors thrown at runtime, sometimes after multiple hours of training.

\etitle{Pattern 3:} Building production-grade applications making use of machine learning pipelines is a laborious task. As a first step, one needs to formalize the prediction task, and choose the components:  feature construction and transformations, the training algorithm,  hyper parameters and their tuning. Once the pipeline is developed and successfully trained, it must be integrated into the application and shipped in production. 
This process is usually performed by a different engineer (or team) than the one building the model, and a significant rewriting is often required because of various runtime constraints (e.g., a different hardware or software platform, constraints on pipeline size or prediction latency/throughput). Such rewriting is often done for particular applications, resulting in custom solutions: a process not sustainable at \msft scale.

Solving the issues revealed by the above  patterns requires rethinking the ML framework for pipeline composition.  Key requirements for it can be summarized as follows:
\vspace{-1ex}
\begin{enumerate}
\item \emph{Unification}: \tool must act as a unifying framework that can host a variety of models and components (with related idiosyncrasies). Once ML pipelines are trained, the same pipeline must be deployable into any production environment (from data centers to IoT devices) with close to zero engineering cost.
In the last decade 100s of products and services have employed \tool, validating its success as a unifying platform. 


\item \emph{Extensibility}: Data scientists are interested in experimenting with different models and features with the goal of obtaining the best accuracy. Therefore, it should be possible to add 
new components and algorithms with minimal reasonable effort via a general API that supports a variety of data types and formats. 
Since its inception, 
\tool has been extended with many components. In fact, a large fraction of the now more than 120 built-in operators started life as extensions shared between data scientists.

\item \emph{Scalability and Performance}: \tool must be scalable and allow maximum hardware utilization---i.e., be fast and provide high throughput. Because production-grade datasets are  often very large and do not fit in RAM, scalability implies the ability to run pipelines in out-of-memory mode, with data paged in and processed incrementally. 
As we show in the experiment section, \tool achieves good scalability and performance (up to several orders-of-magnitude) when compared to other publicly available toolkits.

\end{enumerate}


\stitle{\tool: Overview.}
\tool is a {.NET} machine learning library that allows developers to build complex machine learning pipelines, evaluate them, and then utilize them directly for prediction. 
Pipelines are often composed of multiple transformation steps that featurize and transform the raw input data, followed by one or more ML models that can be stacked or form ensembles. Note that ``pipeline'' is a bit of a misnomer, as they, in fact, are Direct Acyclic Graphs (DAGs) of operators. We next illustrate how these tasks can be accomplished in \tool on a short example~\footnote{This and other examples can be accessed at \url{https://github.com/dotnet/machinelearning-samples}.}; we will also exploit this example to introduce the main concepts in \tool.

\begin{figure}[h]
\vspace{-2ex}
\lstset{numbersep=5pt, xleftmargin=10pt,   framexleftmargin=10pt}
\begin{lstlisting}
var ml = new MLContext();
var data = ml.Data.LoadFromTextFile<SentimentData>(trainingDatasetPath);
var pipeline = ml.Transforms.Text.FeaturizeText("Features","Text")
    .Append(ml.BinaryClassification.Trainers.FastTree());
\end{lstlisting}
\vspace{-5ex}
\caption{A pipeline for text analysis whereby input sentences are classified according to the expressed sentiment.}
\label{fig:sa}
\vspace{-1ex}
\end{figure}

Figure~\ref{fig:sa} introduces a Sentiment Analysis pipeline (SA).  
The first item required for building a pipeline is the \emph{MLContext} (line 1): the entry-point for accessing \tool features. In line 2, a \emph{loader} is used to indicate how to read the input training data.  In the example pipeline, the input schema (\texttt{SentimentData}) is specified explicitly, but in other situations (e.g., CSV files with headers) schemas can be automatically inferred by the loader.  Loaders generate a \idataview object, which is the core data abstraction of \tool. \idataview provides a fully schematized non-materialized view of the data, and gets subsequently transformed by pipeline components.  

The second step is feature extraction from the input \texttt{Text} column  (line 3).  To achieve this, we use the \texttt{FeaturizeText} \emph{transform}. Transforms are the main \tool operators for manipulating data. Transforms accept a \idataview as input and produce another \idataview. 
\texttt{FeaturizeText} is actually a complex transform built off a composition of nine base transforms that perform common tasks for feature extraction from natural text. Specifically, the input text is first normalized and tokenized. For each token, both char- and word-based ngrams are extracted and translated into vectors of numerical values. These vectors are subsequently normalized 
and concatenated to form the final \texttt{Features} column. 
Some of the above transforms (e.g., normalizer) are \emph{trainable}: i.e., before producing an output \idataview they are required to scan the whole dataset to determine internal parameters (e.g., scalers).

Subsequently, in line 4 we apply a \emph{learner} (i.e., a trainable model) to the pipeline---in this case, a binary classifier called FastTree: an implementation of the MART gradient boosting algorithm~\cite{Friedman00}. 
Once the pipeline is assembled, we can train it by calling the \texttt{Fit} method on the pipeline object with the expected output prediction type (Figure~\ref{fig:sa-training}).
\tool evaluation is \emph{lazy}: no computation is actually run until the \texttt{Fit} method (or other methods triggering pipeline execution) is called. This allows \tool to (1) properly validate that the pipeline is well-formed before computation; and (2) deliver state of the art performance by devising efficient execution plans.

\lstset{frame=tb,
  language=Scala,
  aboveskip=5mm,
  belowskip=0mm,
  showstringspaces=false,
  columns=flexible,
  basicstyle={\scriptsize\ttfamily},
  numberstyle=\tiny\color{gray},
  keywordstyle=\color{blue},
  commentstyle=\color{dkgreen},
  stringstyle=\color{mauve},
  breaklines=true,
  breakatwhitespace=true,
  tabsize=3,
  numbers=none,
  xleftmargin=1em,
  framexleftmargin=1em,
  abovecaptionskip=5mm,
  captionpos=b
}
\begin{figure}[h]
\vspace{-2.5ex}
\begin{lstlisting}
var model = pipeline.Fit(data);
\end{lstlisting}
\vspace{-2ex}
\caption{Training of the sentiment analysis pipeline. Up to this point no execution is actually triggered.}
\label{fig:sa-training}
\vspace{-1ex}
\end{figure}

Once a pipeline is trained, a model object containing all training information is created. The model can be saved to a file (in this case, the information of all trained operators
as well as the pipeline structure are serialized into a compressed file), or evaluated against a test dataset (Figure~\ref{fig:sa-eval}) or directly used for prediction serving (Figure~\ref{fig:sa-serving}).
To evaluate model performance, \tool provides specific components called \emph{evaluators}. Evaluators accept as input a \idataview upon which a model has been previously applied, and produce a set of metrics. 
In the specific case of the evalutor used in Figure~\ref{fig:sa-eval}, relevant metrics are those used for binary classifiers, such as accuracy, Area Under the Curve (AUC), log-loss, etc.

\begin{figure}[h]
\vspace{-3ex}
\begin{lstlisting}
var output = model.Transform(testData);
var metrics = mlContext.BinaryClassification.Evaluate(output);
\end{lstlisting}
\vspace{-2ex}
\caption{Evaluating mode accuracy using a test dataset.}
\label{fig:sa-eval}
\vspace{-1ex}
\end{figure}

Finally, serving the model for prediction is achieved by first creating a \emph{PredictionEngine}, and then calling the \texttt{Predict} method with a list of \texttt{SentimentData} objects.  Predictions can be served natively in any OS (e.g., Linux, Windows, Android, macOS) or device supported by the .NET Core framework.

\begin{figure}[h]
\vspace{-3ex}
\begin{lstlisting}
var engine = ml.Model
    .CreatePredictionEngine<SentimentData, SentimentPrediction>(model);
var predictions = engine.Predict(PredictionData);
\end{lstlisting}
\vspace{-2ex}
\caption{Serving predictions using the trained model.}
\label{fig:sa-serving}
\vspace{-3ex}
\end{figure}

%% file: design.tex
\section{System Design and Abstractions}
\label{sec-design}

In order to address the requirements listed in Section~\ref{sec-motivation}, \tool borrows ideas from the database community.
\tool's main abstraction is called \idataview (Section~\ref{sec-idataview}). 
Similarly to (intensional) database relations, the \idataview abstraction provides compositional processing of schematized data, but specializes it for machine learning pipelines. 
The \idataview abstraction is generic and supports both primitive operators as well as the composition of multiple operators to achieve higher-level semantics such as the \texttt{FeaturizeText} transform of Figure~\ref{fig:sa} (Section~\ref{sec-composition}).
Under the hood, operators implementing the \idataview interface are able to gracefully and efficiently handle high-dimensional and large datasets thanks to \emph{cursoring} (Section~\ref{sec-cursoring}) which resembles the well-known iterator model of databases~\cite{iterator-model}.

\vspace{-1ex}
\subsection{The \idataview Abstraction}
\label{sec-idataview}

In relational databases, the term 
\emph{view} typically indicates the result of a query on one or more tables (base relations) or views, and is generally immutable. Views (and tables) are defined over a \emph{schema} which expresses a sequence of \emph{columns names} with related \emph{types}. 
The semantics of the schema is such that each data row outputs of a view must conform to its schema.
%
%
%
Views have interesting properties which differentiate them from tables and make them appropriate abstractions for machine learning: (1) views are \emph{composable}---new views are formed by applying transformations (queries) over other views; (2) 
views are \emph{virtual}, i.e., they can be lazily computed on demand from other views or tables without having to materialize any partial results; and
(3) since a view does not contain values, but merely computes values from its source views, it is \emph{immutable} and \emph{deterministic}: the same exact computation applied over the same input data always produces the same result.
Immutability and deterministic computation (note that several other data processing systems such as Apache Spark~\cite{spark} employ the same assumptions) enable transparent data caching (for speeding up iterative computations such as ML algorithms) and safe parallel execution.
\idataview inherits the aforementioned database view properties, namely: schematization, composability, lazy evaluation, immutability, and deterministic execution.

\stitle{Schema with Hidden Columns.} Each \idataview carries schema information specifying the name and type of each view's column. 
\idataview schemas are ordered and, 
by design, multiple columns can share the same name, in which case, one of the columns \emph{hides} the others: referencing a column by name always maps to the latest column with that name.
Hidden columns exist because of immutability and can be used for debugging purposes: 
having all partial computations stored as hidden columns allows the inspection of the provenance of each data transformation.
\eat{Figure~\ref{} depicts this situation in which 2 \idataview $\!$s generate a column with the same name (\texttt{Features}). 
From a user's perspective (which is entirely based on column names), the \texttt{Features} column was ``modified'' by the last computation, but the original values are available downstream via the hidden columns.
}
Indeed, hidden columns are never fully materialized in memory (unless explicitly required) therefore their resource cost is minimal.
 
\stitle{High Dimensional Data Support with Vector Types.} While the \idataview schema system supports an arbitrary number of columns, like most schematized data systems, it is designed for a modest number of columns, typically, limited to a few hundred. 
Machine learning and advanced analytics applications often involve high-dimensional data. For example, common techniques for learning from text uses bag-of-words (e.g., \texttt{FeaturizeText}), one-hot encoding or hashing variations to represent non-numerical data. These techniques typically generate an enormous number of features. 
Representing each feature as an individual column is far from ideal, both from the perspective of how the user interacts with the information and how the information is managed in the schematized system. The \idataview solution is to represent each set of features as a single \emph{vector} column.
A vector type specifies an item type and optional dimensionality information. The item type must be a primitive, non-vector, type. The optional dimensionality information specifies the number of items in the corresponding vector values.
When the size is unspecified, the vector type is variable-length. For example, the \texttt{TextTokenizer} transform (contained in \texttt{FeaturizeText}) maps a text value to a sequence of individual terms. This transformation naturally produces variable-length vectors of text. 
Conversely, fixed-size vector columns are used, for example, to represent a range of column from an input dataset.
\vspace{-1ex}

\subsection{Composing Computations using \idataview}
\label{sec-composition}

\tool includes several standard operators and the ability to compose them using the \idataview abstraction to produce efficient machine learning pipelines. 
\emph{Transform} is the main operator class: transforms are applied to a \idataview to produce a derived \idataview and are used to prepare data for training, testing, or prediction serving.
\emph{Learners} are machine learning algorithms that are trained on data (eventually coming from some transform) and produce predictive models. 
\emph{Evaluators} take scored test datasets and produced metrics such as precision, recall, F1, AUC, etc.
Finally, \emph{Loaders} are used to represent data sources as a \idataview, while \emph{Savers} serialize \idataviews to a form that can be read by a loader.
We now details some of the above concepts.

\stitle{Transforms.}
Transforms take a \idataview as input and produce a \idataview as output. Many transforms simply ``add'' one or more computed columns to their input schema. More precisely, their output schema includes all the columns of the input schema, plus some additional columns, whose values are computed starting from some of the input columns. It is common for an added column to have the same name as an input column, in which case, the added column hides the input column, as we have previously described. 
Multiple primitive transforms may be applied to achieve higher-level semantics: for example, the \texttt{FeaturizeText} transform of Figure~\ref{fig:sa} is the composition of 9 primitive transforms. 

\stitle{Trainable Transforms.}
While many transforms simply map input data values to output by applying some pre-defined computation logic (e.g., \texttt{Concat}), other transforms require ``training'', i.e., their precise behavior is determined automatically from the input training data. For example, normalizers and dictionary-based mappers translating input values into numerical values (used in \texttt{FeaturizeText}) build their state from training data. 
Given a pipeline, a call to \texttt{Train} triggers the execution of all trainable transforms ( as well as learners) in topological order.
When a transform (learner) is trained, it produces a \idataview representing the computation up to that point in the pipeline: the \idataview can then be used by downstream operators.  
Once trained and later saved, the state of a trained transform is serialized such that, once loaded back the transform is not retrained. 

\stitle{Learners.} Similarly to trainable transforms, learners are machine learning algorithms that take \idataview as input and produce ``models": transforms that can be applied over input \idataviews and produce predictions. \tool supports learners for \emph{binary classification}, \emph{regression}, \emph{multi-class classification}, \emph{ranking}, \emph{clustering}, \emph{anomaly detection}, \emph{recommendation} and \emph{sequence prediction} tasks.



\subsection{Cursoring over Data}
\label{sec-cursoring}

\tool uses \idataview as a representation of a computation over data. 
Access to the actual data is provided through the concept of \emph{row cursor}.
While in databases queries are compiled into a chain of operators, each of them implementing an iterator-based interface, in \tool, ML pipelines are compiled into chains of \idataviews where data is accessed through cursoring.  
A row cursor is a movable window over a sequence of data rows coming either from the input dataset or from the result of the computation represented by another \idataview. 
The row cursor provides the column values for the current row, and, as iterators, can only be advanced forward (no backtracking is allowed).


\stitle{Columnar Computation.} 
In data processing systems, it is common for a down-stream operator to only require a small subset of the information produced by the upstream pipeline. 
For example, databases have columnar storage layouts to avoid access to unnecessary columns~\cite{cstore}. 
This is even more so in machine learning pipelines where featurizers and ML models often work on one column at a time.
For instance, \texttt{FeaturizeText} needs to build a dictionary of all terms used in a text column, while it does not need to iterate over any other columns. 
\tool provides columnar-style computation model through the notion of \emph{active columns} in row cursors. 
Active columns are set when a cursor is initialized: the cursor then enforces the contract that only the computation or data movement necessary to provide the values for the active columns are performed. 

\stitle{Pull-base Model, Streaming Data.}
\tool runtime performance are proportional to data movements and computations required to scan the data rows. 
As iterators in database, cursors are pull-based: after an initial setup phase (where for example active columns are specified) cursors do not access any data, unless explicitly asked to. 
This strategy allows \tool to perform at each time only the computation and data movements needed to materialize the requested rows (and column values within a row).
For large data scenarios, this is of paramount importance because it allows efficient streaming of data directly from disk, without having to rely on the assumption that working sets fit into main memory.
Indeed, when the data is known to fit in memory, caching provides better performance for iterative computations.

%% file: implementation.tex
\section{System Implementation}
\label{sec-implementation}

\tool is the solution \msft developed for the problem of empowering developers with a machine learning framework to author, test and deploy ML pipelines.
As introduced in Section~\ref{sec-motivation}, \tool is implemented with the goal of providing a tool that is easy to use, scalable over large datasets while providing good performance, and able to unify under a single API data transformations, featurizers, and state of the art machine learning models. 
In its current implementation, \tool comprises 2773K lines of C\# code, and about 74K lines of C++ code, the latter used mostly for high-performance linear algebra operations employing SIMD instructions. 
\tool supports more then 80 featurizers and 40 machine learning models.



\subsection{Writing Machine Learning Pipelines in \tool}
\label{sec-apis}

\tool comes with several APIs, all covering the different use cases we observed during the years at \msft. All APIs eventually are compiled into the typed learning pipeline API with generics shown in the SA example of Section~\ref{sec-motivation}.
Beyond the typed API with generics, \tool supports a (1) command line / scripting API enabling data scientists to easily experiment with several pipelines; (2) a Graphical User Interface for users less familiar with coding; 
and (3) an \emph{Entry Point} (EP) API allowing to execute and code-generate APIs in different languages (e.g., Scala and Python). 
Due to space constraint, next we only detail the EP API and one of its applications, namely NimbusML~\cite{nimbus}: a Python API mirroring \skl pipeline API.

\stitle{Entry Points and Graph Runner.}
The recommended way of interacting with \tool through other, non-.NET, programming languages is by composing, and exchanging \emph{entry point graphs}.
An EP is a JSON representation of a \tool operator.
EPs descriptions are grouped into a manifest file: a JSON object that documents and defines the structure of any available EP. 
The operator manifest is code-generated by scanning the \tool assemblies through reflection and searching for specific types and class annotations of operators.
Using the EP API, pipelines are represented as graphs of EP operators and input/output relationships which are all serialized into a JSON file.
EP graphs are parsed in \tool by the \emph{graph runner} component which generates and directly executes the related pipeline.
Non-.NET APIs can be automatically generated starting from the manifest file so that there is no need to write and maintain additional APIs.

\begin{figure}[h]
 \includegraphics[width=0.48\textwidth]{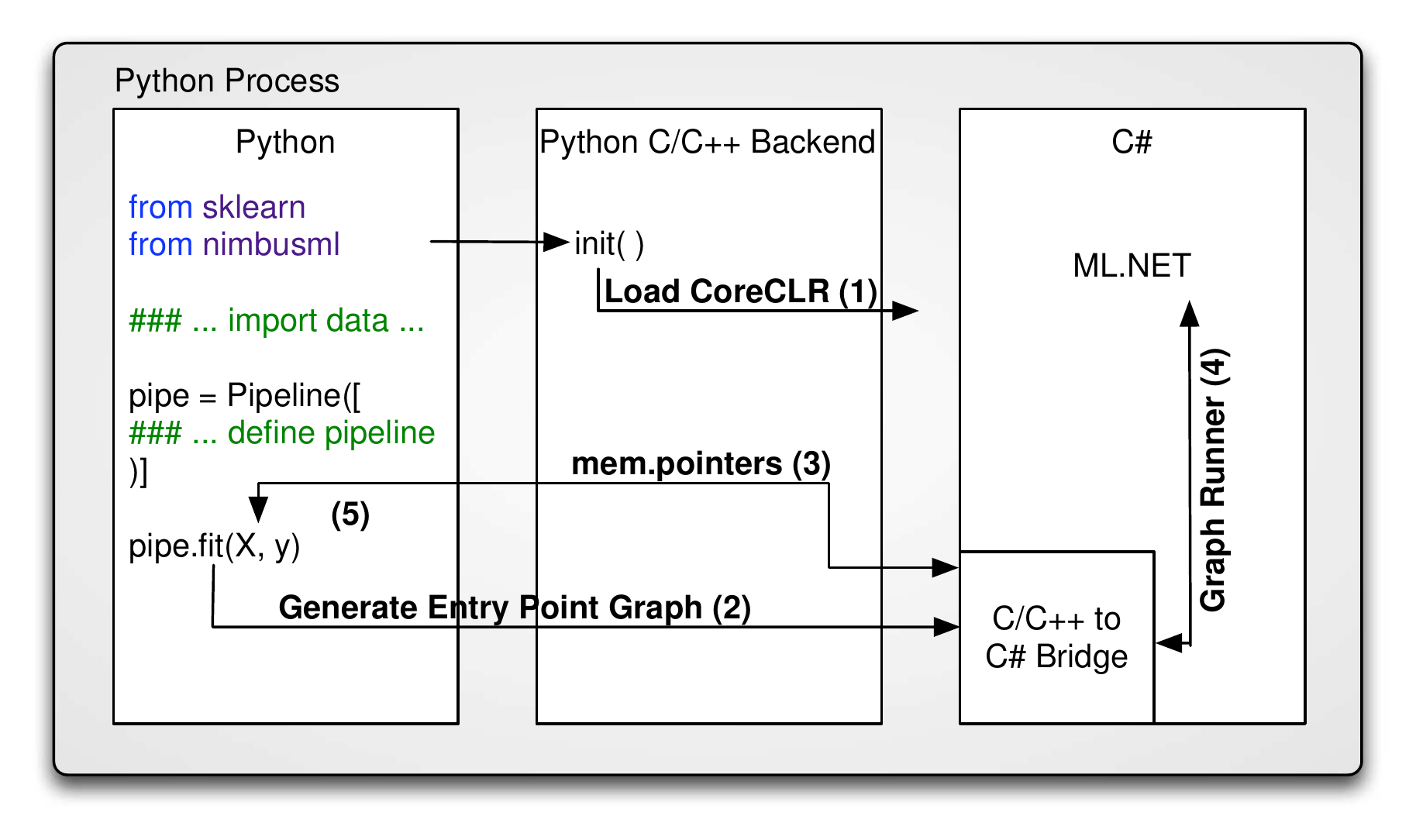}
     \vspace{-6ex}
  	\caption{Architecture and execution flow of NimbusML. When NimbusML is imported, a CoreCLR instance with \tool is initialized (1). When a pipeline is submitted for execution, the entry point graph of NimbusML operators is extracted and submitted for execution (2) together with pointers to the datasets (3) if residing in memory. The graph runner executes the graph (4) and returns the result to the Python program as a data frame object (5).}
    \label{fig:pysys}
    \vspace{-2ex}
\end{figure}

\stitle{NimbusML and Interoperability with \skl.}
NimbusML is \tool's Python API mirroring \skl interface (NimbusML operators are actually subclasses of \skl components) and taking advantage of the EP API functionalities. 
Furthermore, data scientists can start with a \skl pipeline and swap \skl transformations or algorithms with \tool's ones to achieve better scalability and accuracy.
To achieve this level of interoperability, however, data residing in \skl needs to be accessed from \tool (and vice-versa), but the EP API does not provide such functionality.
To obtain such behavior in an efficient  and scalable way, when NimbusML is imported into a \skl project, a .NET Core runtime as well as an instance of \tool are spawn within the same Python process.
When a user triggers the execution of a pipeline, the call is intercepted on the Python side and an EP graph is generated, and submitted to the graph runner component of the \tool instance.
If the data resides in memory in \skl as a Pandas data frame or a numpy array, the C++ reference of the data is passed to the graph runner through C\#/C++ interop and wrapped around a \idataview, which is then used as input for the pipeline.
We used Boost.Python~\cite{boot-python} as helper library to access the references of data residing in \skl. 
If the data resides on disk, a \idataview is instead directly used.
Figure~\ref{fig:pysys} depicts the architecture of NimbusML.

\vspace{-2ex}

\subsection{Pipeline Execution}
Independently from which API is used by the developer or the task to execute (training, testing or prediction), \tool eventually runs a learning pipeline. 
As we will describe shortly, thanks to lazy evaluation, immutability and the Just In Time (JIT) compiler provided by the .NET runtime, \tool is able to generate highly efficient computations.
Internally, transforms consume \idataview columns as input and produce one (or more) \idataview columns as output. Columns are immutable whereby multiple downstream operators can safely consume the same input without triggering any re-execution. 
Trainable transforms and learners, instead, need to be trained before generating the related output \idataview column(s).
Therefore, when a pipeline is submitted for execution (e.g., by calling \texttt{Train}), each trainable transform / learner is trained in topological order.
For each of them, a one-time initialization cost is payed to analyze the cursors in the pipeline, e.g., each cursor checks the active columns and the expected input type(s).

\stitle{CPU Efficiency.}
Output of the initialization process at each \texttt{Data} \texttt{View}'s cursor is a lambda function, named \at{getter}, condensing the logic of the operator into one single call. 
Each \at{getter} in turn triggers the generation of the \at{getter} function of the upstream cursor until a data source is found (e.g., a cached \idataview or input data).
When all \at{getters} are initialized, each upstream \at{getter} function is used in the downstream one, so that, from the outer cursor perspective, computation is represented as a chain of lambda function calls.
Once the initialization process is complete, the cursor iterates over the input data and executes the training (or prediction) logic by calling its \at{getter}.
At execution time, the chain of \at{getter} are JIT-compiled by the .NET runtime to form a unique, highly efficient function executing the whole pipeline (up to that point) on a single call. 
The process is repeated until no trainable operator is left in the pipeline.

\stitle{Memory Efficiency.} 
Cursoring is inherently efficient from a memory allocation perspective. Advancing the cursor to the next row requires no memory allocation. Retrieving primitive column values from a cursor also requires no memory allocation. To retrieve vector column values from a cursor, the caller to the \at{getter} can optionally provide buffers into which the values should be copied. When the provided buffers are sufficiently large, no additional memory allocation is required. When the buffers are not provided or are too small, the cursor allocates buffers of sufficient size to hold the values. This cooperative buffer sharing protocol eliminates the need to allocate separate buffers for each row.

\stitle{Parallel Computation.}
\tool provides 2 possibilities for improving performance through parallel processing: (1) from directly inside the algorithm; and (2) using parallel cursoring.
The former case is strictly related to the algorithm implementation.
In the latter case, a transform requires a \emph{cursor set} from its input \idataview.
Cursors sets are propagated upstream until a data source is found: at this point cursor set are mapped into available threads, and data is collaboratively scanned.
From a callers perspective, cursor sets return a consolidated, unique, cursor, although, from an execution perspective, cursor's data scan is split into concurrent threads.

\subsection{Learning Tasks in \tool}
We have surveyed the top learners by number of unique users within \msft. 
We will here subdivide the usage by what appears to be most popular.~\footnote{
Note that using unique users to asses popularity is indeed wrong: just because a learner is not popular does not mean that its support is not strategically important.}

\stitle{Gradient Boosting Trees.}
The most popular \emph{single} learner is FastTree: a gradient boosting algorithm over trees. FastTree uses an algorithm that was originally engineered for web-page ranking, but was later adapted to other tasks---and in fact the ranking task, while still having thousands of unique users within \msft, is comparatively much less popular than the more classical tasks of binary classification and regression. This learner requires a representation of the dataset in memory to function.
Interestingly, a random-forest algorithm based on the same underlying code sees only a small fraction of the usage of the boosting-based interface.
As point of interest, a faster implementation of the same basic algorithm called LightGBM~\cite{lgbm} was introduced few years ago, and is gaining in popularity. However, usage of LightGBM still remains a fraction of the original algorithm, possibly for reasons of inertia.

\stitle{Linear Learners.}
While the most popular learner is based on boosted decision trees, one could argue that collectively linear learners see more use. 
Linear learners in contrast to the tree based algorithm do work well over streaming data.
The most popular linear learners are basic implementations of such familiar algorithms like \emph{averaged perceptron}, \emph{online gradient descent}, \emph{stochastic gradient descent}. These scale well and are quite simple to use, though they lack the sophistication of other methods.
Following this ``basic set'' in popularity are a set of linear learners based on OWL-QN~\cite{owlqn}, a variant of L-BFGS capable of L1 regularization, and thus learning sparse linear prediction rules. These algorithms have some advantages, but because these algorithms on the whole require more passes over the dataset to converge to a good answer compared to the earlier stochastic methods, they are less popular.
Even less popular still is an SDCA-based~\cite{sdca} algorithm, as well as a linear Pegasos SVM trainer~\cite{pegaso}. Each still has in excess of a thousand users, but this is still considerably less than the other algorithms.

\stitle{Other Learners.}
Compared to these supervised tasks, unsupervised tasks like clustering and anomaly detection are definitely part of the long tail, each having perhaps only hundreds of unique users. Even more obscure tasks like sequence classification and recommendation, despite supporting quite important products, seem to have only a few unique users. 
Readers may note neural networks and deep learning as a very conspicuous omission. We do internally have a neural network that sees considerable use, but
in the open source version we instead provide an interface to other, already available, neural network frameworks.

%% file: experiments.tex
\section{Experimental Evaluation}
\label{sec-experiments}


In this section we compare \tool against \sk and H2O.
For \tool we employ its regular learning pipeline API (ML.NET) and the Python bindings through NimbusML. For NimbusML we test both reading from Pandas' Data Frames (NimbusML-DF) and the streaming API (NimbusML-DV) of \idataview which allows to directly stream data from disk.
We tried as much as we could to use the same data transforms and ML models across different toolkits in order to measure the performance of the frameworks and not of the data scientist.
For the same reason, for each pipeline we use the default parameters.
We report the total runtime (training plus testing), AUC for the classification problems, and Root Mean Square (RMS) for the regression one.
To examine the scale-out performance of the frameworks, in the first set of experiments we train the pipelines over 0.1\%, 1\%, 10\% and 100\% of samples of the training data over all accessible cores.
Finally, Section~\ref{sec:scaleup} contains a scale-up experiment where we measure how the performance change as we change the number of used cores. For this final set of experiments we only compare ML.NET/NimbusML and H2O because \skl is only able to use a single core.

All the experiments are run three times and the minimum runtime and the average accuracy are reported. Further information about the experiments are reported into the Reproducibility Section attached to the end of the paper.

\vspace{1ex}
\stitle{Configuration.}
All the experiments in the paper are carried out on Azure on a D13v2
VM
with 56GB of RAM, 112 GB of Local SSD and a single Intel(R) Xeon(R) @ 2.40GHz processor. We used ML.NET version 0.1, NimbusML version 0.6, \skl version 0.19.1 and H2O version 3.20.0.7.

\vspace{1ex}
\stitle{Scenarios.}
In our evaluation we train models for four different scenarios. In the first scenario we aim at predicting the click through rate for an online advertisement.
In the second scenario we train a model to predict the sentiment class for e-commerce customer reviews.
In the third scenario we predict the delay for scheduled flights according to historical records. 
(Note that the first two are classification problems, while the third one is a  regression problem.) 
In these first three scenario we allow the systems to uses all the available processing resources.
Conversely, in the last scenario we report the performance as we scale-up the number of available cores. 
For this scenario we chose two different test situations: one where the dataset is a small sample (1\%), and one where the dataset is large. In this way we can evaluate the difference scale-up performance.

\vspace{1ex}
\stitle{Datasets.}
For each scenario we use a different dataset. 
For the first scenario we use the Criteo dataset~\cite{criteo}.
The full training dataset includes around 45 million records, and the size of the training file is around 10GB. Among the 39 features, 14 are numeric while the remaining are categorical.
In the set of experiments for the second scenario we employed the Amazon Review dataset~\cite{amazon}.
In this case the full training dataset includes around 17 million records, and the size of the training file is around 9GB. For this scenario we only use one text column as the input feature.
Finally, for the third scenario we use the Flight Delay dataset~\cite{flightDelay}. 
The training dataset includes around 1 million records, the size of the training file is around 1GB and each record contains 631 columns.


\subsection{Criteo}
\begin{figure}[t!]
\hspace{-2ex}
\includegraphics[width=1.05\columnwidth]{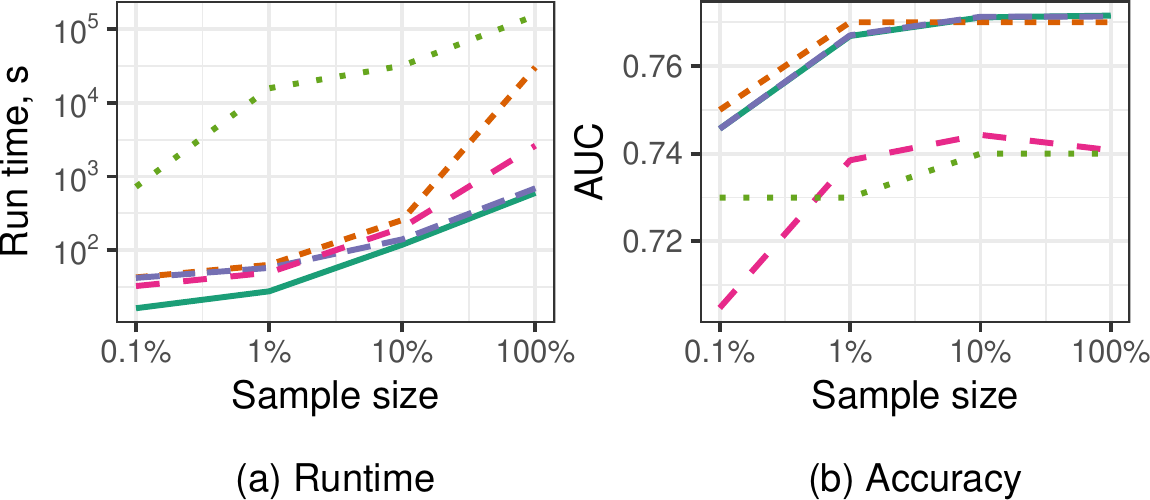}

\vspace{-32ex}
\includegraphics[width=1.05\columnwidth]{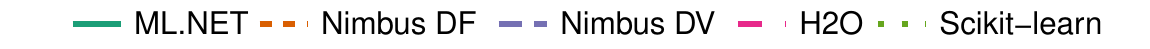}
\vspace{24ex}
\caption{Experimental results for the Criteo dataset.}
\label{fig:rt}
\vspace{2ex}
\end{figure}

In the first scenario we build a pipeline which (1) fills in the missing values in the numerical columns of the dataset; (2) encodes a categorical columns into a numeric matrix using a hash function; and (3) applies a gradient boosting classifier (LightGBM for \tool). 
\eat{
\begin{enumerate}
	\item \textit{Missing\_value.Hander/Imputer}
	\item \textit{OneHotHashVectorizer/FeatureHasher}
	\item \textit{LightGbmClassifier/GradientBoostingClassifier}
\end{enumerate}
The \textit{missing\_value.Hander/Imputer} fills in the missing values in the dataset for numeric columns. The \textit{OneHotHashVectoizer/FeatureHasher} converts categorical columns to a numeric matrix using a hash function. In Sklearn pipelines, as we need to apply different transforms onto different columns and utilize all the outputs as the features for the predictor, we used Sklearn \textit{FeatureUnion} function in the pipeline. 
}
Figure~\ref{fig:rt}a shows the total runtime (including training and testing), while Figure~\ref{fig:rt}b depicts the AUC on the test dataset.
As we can see, ML.NET has the best performance, while NimbusML-DV ranks second. 
Both NimbusML-DF and H2O show good runtime performance, especially for smaller datasets. In this experiment \skl has the worst running time: with the full training dataset, ML.NET and NimbusML-DV train in around 10 minutes while \skl takes more than 2 days.
Regarding the accuracy, we can notice that the results from NimbusML-DV/NimbusML-DF and ML.NET are very similar and all of them dominate \skl/H2O by a large margin.
This is mainly due to the superiority of LightGBM versus the gradient boosting algorithm used in the latters.

\begin{figure}[t!]
\centering
\hspace{-3ex}
\includegraphics[width=1.05\columnwidth]{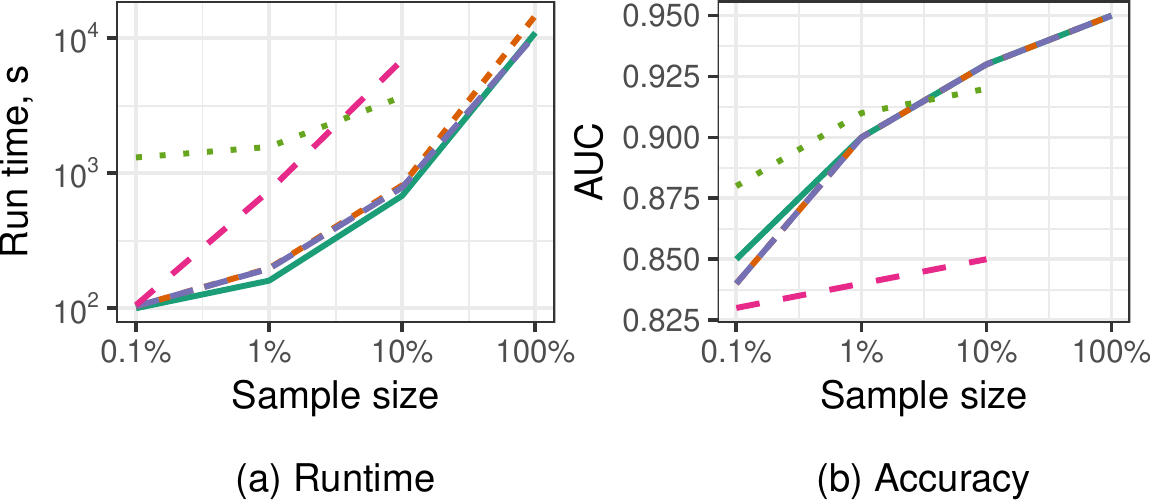}

\vspace{-32ex}
\includegraphics[width=1.05\columnwidth]{legend.pdf}
\vspace{24ex}
\caption{Experimental results for the Amazon dataset.}
\label{fig:am}
\vspace{2ex}
\end{figure}

\begin{figure}[t!]
\centering
\hspace{-3ex}
\includegraphics[width=1.05\columnwidth]{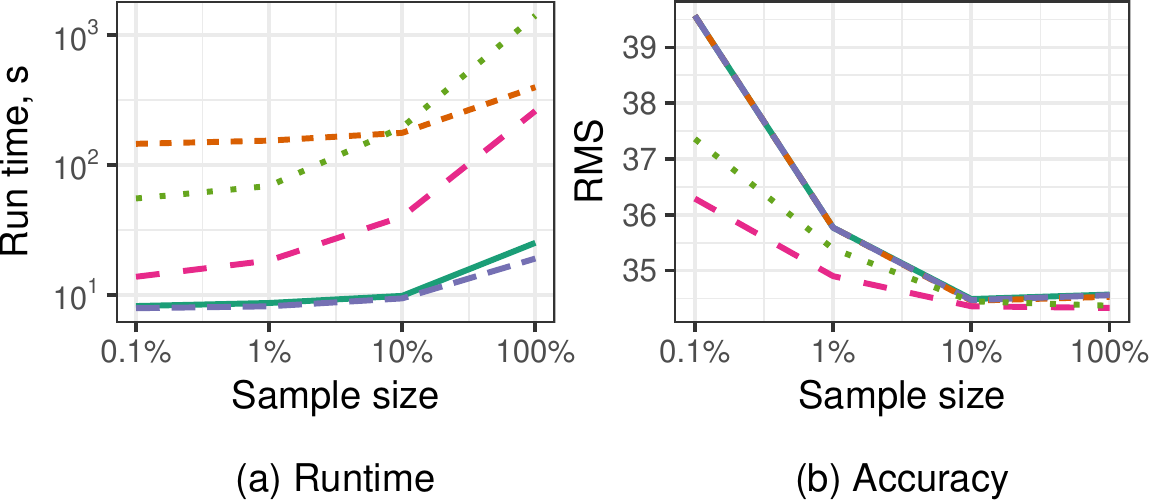}

\vspace{-32ex}
\includegraphics[width=1.05\columnwidth]{legend.pdf}
\vspace{24ex}
\caption{Experimental results for the Flight Delay dataset.}
\label{fig:fd}
\vspace{-2ex}
\end{figure}

\vspace{-2ex}
\subsection{Amazon}
In this scenario the pipeline first featurizes the text column of the dataset and then applies a linear classifier. 
For the featurization part, we used the \texttt{FeaturizeText} transform in \tool and the \texttt{TfidfVectorizer} in \skl.
The \texttt{FeaturizeText/Tfidf} \texttt{Vectorizer} extracts numeric features from the input corps by producing a matrix of token ngrams counts. 
H2O implements the Skip-Gram word2vec model~\cite{word2vec} as the only text featurizer.
\eat{
For \skl we used the , while we employed Word2Vec for H2O.
For the classifier, we used the Average Perceptron in \tool, the regular linear cla

\begin{enumerate}
	\item \textit{NGramFeaturizer/TfidfVectorizer with 1-, 2-word-gram and 1-, 2-, 3-char-gram}, or \textit{Word2Vec} transform for H2O
	\item \textit{PerceptronClassifier/LinearClassifier}
\end{enumerate}
The \textit{NGramFeaturizer/TfidfVecotizer} extracts numeric features for the input corps by producing a matrix of token ngrams counts, e.g. number of times the word ``a" shows up for uni-gram or number of times the phrase ``a word" for bi-gram, etc. Then the counts were normalized according to the total frequency. H2O implements the Skip-Gram word2vec model as the text featurizer. The representation of words were trained through the input corps based on neural nets. 
}
In both the classical approach based on ngrams and the neural network approach, text featurization is a heavy operation.
In fact, both Sklearn and H2O throw overflow/memory errors when training with the full dataset because of the large vocabulary sizes.
Therefore, no results are reported for Sklearn and H2O, trained with the full dataset.
As we can see from Figure~\ref{fig:am}a, \tool is able to complete all experiments, and all versions (ML.NET, NimbusML-DF, NimbusML-DV) show similar runtime performance.
Interestingly enough, NimbusML-DF is able to complete over the full dataset, while \skl is not. This is because, under the hood, NimbusML-DF uses \idataview to execute the pipeline in a streaming fashion, whereas \skl materializes partial results in data frames.
Regarding the measured accuracy (Figure~\ref{fig:am}b), Sklearn shows the highest AUC with 0.1\% of the dataset, likely due to the different initial settings for the algorithm, but for the remaining data points \tool performs better.

\subsection{Flight Delay}
For this dataset we pre-process all the feature columns into numeric values and we compare the performance of a single operator in ML.NET/NimbusML-DV/NimbusML-DF versus Sklearn/H2O without a pipeline (LightGBM vs gradient boosting).
The results for speed and accuracy are reported in Figure~\ref{fig:fd}. Since this is a regression problem, we report the RMS on the test set.
Interestingly, for this dataset NimbusML-DF runtime is considerably worst than ML.NET and NimbusML-DV, and even worst than \skl for small samples.
We found that this is due to the fact that since we are applying the algorithm directly over the input data frame, and the algorithm requires several passes over the data, the overhead of accessing the data residing in C/C++ dominates the runtime.
Additionally we found that the accuracy of ML.NET, especially for smaller samples, is worst than both H2O and \skl.
By examining the execution, we found that with a boosting tree trained with small subset of the data, a simpler tree from Sklearn predicts better than LightGMB in ML.NET. However, with 0.1\% of the dataset, those models are trained with 1000 samples and over 600 features. Therefore models can be easily overfitted. Models from ML.NET converge faster (with much smaller error for the training set, i.e. RMS = 18 for ML.NET and 28 for Sklearn) and are overfitted. In this specific case, Sklearn/H2O models have better performance over the test set as they are less overfitted.
For large training sets, all systems converge to approximately the same accuracy, although ML.NET is more than 10$\times$ faster than \skl and H2O.

\subsection{Scale-up Experiments}
\label{sec:scaleup}

\begin{figure}[t!]
\centering
\hspace{-3ex}
\includegraphics[width=1.0\columnwidth]{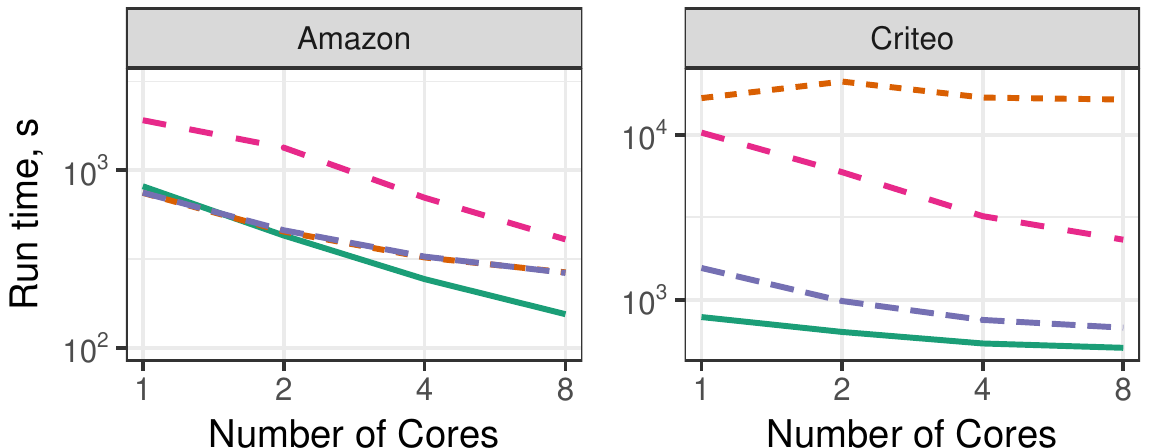}

\vspace{0.5ex}
\includegraphics[width=1.0\columnwidth]{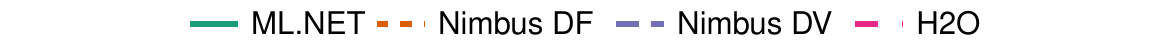}
\vspace{-6ex}
\caption{Scale-up experiments.}
\label{fig:up}
\vspace{-5ex}
\end{figure}

In Figure~\ref{fig:up} we show how the performance of the  systems change as we increase the number of cores. 
For these experiments we use both a small sample (Amazon 1\%, depicted on the left-hand side of Figure~\ref{fig:up}) and a full dataset (Criteo, right-hand side of Figure~\ref{fig:up}) to compare how the systems perform under different stress situations.
For the Amazon sample, \tool and H20 scale linearly, while Nimbus scalability decreases due to the overheads between components.
On the full Criteo experiments, we see that \tool and Nimbus-DV scalability is less compared to H20, however the latter system is about 10$\times$ slower than the formers.
Nimbus-DF performance do not increase as we increase the number of cores because of the overhead of reading/writing Data Frames.
Recall that \skl is not reported for these experiments because parallel training is not supported and therefore performance do not change.

%% file: lessons.tex
\section{Lessons Learned}
\label{sec-lessons}

The current version of \tool is the result of almost a decade of design choices.
Originally there was no \idataview concept nor pipeline but, instead, all featurizers and models computations were applied over an enumerable of \emph{instances}: a vector of floats (sparse or dense) paired with a label (which was itself a float).
With this design, certain types of model classes such as neural networks, recommender systems or time series were difficult to express. Similarly, there was no notion of intermediate computation nor any abstraction allowing to compose arbitrary sequences of operations.
Because of these limitations, \idataview was introduced.
With \idataview, developers can easily stick together different operations, even if internally each operation can be arbitrarily complex and produce an arbitrarily complex output.

In early versions of \idataview, we explored using .NET's \texttt{IEnume\-rable}. At the beginning this was the most natural choice, but we soon found that such approach was generating too many memory allocations. 
This led to the cursor approach with cooperative buffer sharing.
In the first version of \idataview with cursoring, the implementation did not have any \at{getter} but rather a method \texttt{GetColumnValue<T>(int col, ref T val)}. However, the approach had the following problems: (1) every call had to verify that the column was active; (2)
every call had to verify that \texttt{T} was of the right type; and (3) when this is part of a transform in a pipelines (as they often do) each access would be then be accompanied by a virtual method call to the upstream cursor's \texttt{GetColumnValue}.
In contrast, consider the situation with the lambda functions provided by \at{getters}: (1) the verification of whether the column is active happens exactly once; (2) the verification of types happens exactly once; and (3) rather than every access being passed up through a chain of virtual function calls, only a \at{getter} function is used from the cursor, and every data access is executed directly and JIT-ed. 
The practical result of this is that, for some workloads, the ``getter'' method became an order of magnitude faster. 

%% file: related.tex
\section{Related Work}
\label{sec-related}

\skl has been developed as a machine learning tool for Python and, as such, it mainly targets interactive use cases running over datasets fitting in main memory.
Given its characteristic, \skl has several limitations when it comes to experimenting over Big Data: runtime performance are often inadequate; large datasets and feature sets are not supported; datasets cannot be streamed but instead they can only be accessed in batch from main memory. Finally, multi-core processing is not natively supported because of Python's global interpreter lock (although some work exists~\cite{joblib} trying to solve some of these issues for embarrassingly parallel computations such as cross validation or tree ensemble models).
\tool solves the aforementioned problems thanks to the \idataview abstraction (Section~\ref{sec-idataview}) and several other techniques inspired by database systems.
Nonetheless, \tool provides \sk-like Python bindings through NimbusML (Section~\ref{sec-apis}) such that users already familiar with the former can easily switch to \tool. 

MLLib~\cite{mllib}, Uber's Michelangelo~\cite{michelangelo}, H2O~\cite{h2o} and Salesforce's TransmogrifAI~\cite{trans} are machine learning systems built off \skl limitations. Differently than \skl, but similarly to \tool, MLLib, Michelangelo, H2O and TransmogrifAI are not 
``data science languages'' but enterprise-level environments for building machine learning models for applications.
These systems are all JVM-based and they all provide performance for large dataset mainly through in-memory distributed computation (based on Apache Spark~\cite{spark}).
Conversely, \tool main focus is efficient single machine computation.

In Section~\ref{sec-experiments} we compared against H2O because we deem this framework as the closest to \tool. While \tool uses \idataview, 
H2O employs H2O Data Frames as abstraction of data. Differently than \idataview however, H2O Data Frames are not immutable but ``fluid'', i.e., columns can be added, updated and removed by modifying the base data frame. 
Fluid vectors are compressed so that larger than RAM working sets can be used.
H2O provides several interfaces (R, Python, Scala) and large variety of algorithms.

Other popular machine learning frameworks are TensorFlow~\cite{tensorflow}, PyTorch~\cite{pytorch}, CNTK~\cite{cntk}, MXNet~\cite{mxnet}, Caffe2~\cite{caffe2}. These systems however mostly focus on Deep Neural Network models (DNNs).
If we look both internally at \msft, and at external surveys~\cite{kaggle} we find that DNNs are only part of the story, whereas the great majority of models used in practice by data scientists are still generic machine learning models. We are however studying how to merge the two worlds~\cite{neuralTranslator}.



%% file: conclusions.tex
\vspace{-3ex}
\section{Conclusions}
\label{sec-conclusions}
\vspace{-0.5ex}

Machine learning is rapidly transitioning from a niche field to a core element of modern application development. This raises a number of challenges \msft faced early on. \tool addresses a core set of them: it brings machine learning onto the same technology stack as application development, delivers the scalability needed to work on datasets large and small across a myriad of devices and environments, and, most importantly, allows for complete pipelines to be authored and shared in an efficient manner. These attributes of \tool are not an accident: they have been developed in response to requests and insights from thousands of data scientists at \msft who used it to create hundreds of services and products used by hundreds of millions of people worldwide every day.
\tool~\cite{mldotnet} and NimbusML~\cite{nimbus} are open source and publicly available under the MIT license.

\eat{
The number of machine learning frameworks that has been developed and open sourced in the last few years suggests that ML is finally moving from a science mastered by few to a technology accessible to every developer.
In this paper we introduced \tool, \msft's solution for enabling developers to author and deploy ML models in their intelligent applications.
Thanks to the \idataview abstraction, \tool is able to provide large scale, memory efficient    execution of machine learning pipelines composed by data featurizers and ML models.
\tool's usability is validated by the fact that hundreds of developers within \msft have been using it over the years and in several product.
We experimentally evaluated \tool's state of the art performance against well known ML frameworks such as \skl and H2O.
\tool is open source under the MIT license.
}

%% file: reproducibility.tex
\section*{Reproducibility}
In this section we report additional information and insights regarding the experimental evaluation of Section~\ref{sec-experiments}.
For reproducibility we added the Python scripts (with the employed parameters) we used for ML.NET (i.e., NimbusML with Data Frames), \skl and H2O.

\stitle{Criteo.}
For this scenario we build a pipeline which (1) fills in the missing values in the numerical columns of the dataset; (2) encodes a categorical columns into a numeric matrix using a hash function; and (3) applies a gradient boosting classifier (LightGBM for \tool / NimbusML). 
The training scripts used for NimbusML, \skl,  and H20 are shown in Figure~\ref{fig:criteo-ml}, Figure~\ref{fig:criteo-sci}, and Figure~\ref{fig:criteo-h20}, respectively.

\lstset{frame=tb,
  language=Python,
  aboveskip=3mm,
  belowskip=5mm,
  showstringspaces=false,
  columns=flexible,
  basicstyle={\scriptsize\ttfamily},
  numberstyle=\tiny\color{gray},
  keywordstyle=\color{blue},
  commentstyle=\color{dkgreen},
  stringstyle=\color{mauve},
  breaklines=true,
  breakatwhitespace=true,
  tabsize=3,
  numbers=left,
  xleftmargin=1em,
  framexleftmargin=1em,
}
\lstdefinestyle{fault}{ numbers=none, xleftmargin=1.5em , otherkeywords={ =>,<=, ==, > , ||}}

\begin{figure}[h]
\vspace{-3ex}
\lstset{numbersep=5pt, xleftmargin=10pt,   framexleftmargin=10pt}
\begin{lstlisting}
import pandas as pd
from nimbusml.preprocessing.missing_values import Handler
from nimbusml.feature_extraction.categorical import OneHotHashVectorizer
from nimbusml import Pipeline
from nimbusml.ensemble import LightGbmBinaryClassifier

def LoadData(fileName):
    data = pd.read_csv(fileName, header = None, sep='\t', error_bad_lines=False)
    colnames = [ 'V'+str(x) for x in range(0, data.shape[1])]
    data.columns = colnames
    data.iloc[:,14:] = data.iloc[:,14:].astype(str)
    return data.iloc[:,1:],data.iloc[:,0]  

# Load data
dataTrain, labelTrain = LoadData(train_file)

# Create pipeline
nimbus_ppl = Pipeline([
    Handler(columns = [ 'V'+str(x) for x in range(1, 14)]),
    OneHotHashVectorizer(columns = [ 'V'+str(x) for x in range(14, 40)]),
    LightGbmBinaryClassifier(feature = [ 'V'+str(x) for x in range(1, 40)])])

# Train
nimbus_ppl.fit(dataTrain,labelTrain,parallel = n_thread)
\end{lstlisting}
\vspace{-6ex}
\caption{Criteo in NimbusML with DataFrames.}
\label{fig:criteo-ml}
\vspace{-3ex}
\end{figure}

\begin{figure}[H]
\vspace{-2ex}
\lstset{numbersep=5pt, xleftmargin=10pt,   framexleftmargin=10pt}
\begin{lstlisting}
import pandas as pd
from sklearn.preprocessing import Imputer, FunctionTransformer
from sklearn.feature_extraction import FeatureHasher as FeatureHasher
from sklearn.ensemble import GradientBoostingClassifier as GradientBoostingClassifier
from sklearn.pipeline import Pipeline, make_pipeline, FeatureUnion

# Helpfer funciton for sklearn pipeline to select columns
def CategoricalColumn(X):
    return X[[ 'V'+str(x) for x in range(14, 40)]].astype(str).to_dict(orient = 'records')
def NumericColumn(X):
    return X[[ 'V'+str(x) for x in range(1, 14)]]

def LoadData(fileName):
    data = pd.read_csv(fileName, header = None, sep='\t', error_bad_lines=False)
    colnames = [ 'V'+str(x) for x in range(0, data.shape[1])]
    data.columns = colnames
    data.iloc[:,14:] = data.iloc[:,14:].astype(str)
    return data.iloc[:,1:],data.iloc[:,0]  

# Load data
dataTrain, labelTrain = LoadData(train_file)

# Create pipeline
imp = make_pipeline(FunctionTransformer(NumericColumn, validate=False), Imputer()) 
hasher = make_pipeline(FunctionTransformer(CategoricalColumn, validate=False), FeatureHasher())
sk_ppl = Pipeline([  
   ('union', FeatureUnion(
             transformer_list=[('hasher',hasher), ('imp',imp)]),
   ('tree', GradientBoostingClassifier())]) 
    
# Train
sk_ppl.fit(dataTrain,labelTrain)
\end{lstlisting}
\vspace{-6ex}
\caption{Criteo training pipeline in \skl.}
\label{fig:criteo-sci}
\vspace{-2ex}
\end{figure}

\begin{figure}[h]
\vspace{-2ex}
\lstset{numbersep=5pt, xleftmargin=10pt,   framexleftmargin=10pt}
\begin{lstlisting}
import h2o
from h2o.estimators.gbm import H2OGradientBoostingEstimator

h2o.init(nthreads=n_thread, min_mem_size='20g')

# Load data
dataTrain = h2o.import_file(train_file,header = 0)

# Create pipeline
response = "C1"
features = ["C" + str(x) for x in range(2,41)]
dataTrain[response] = dataTrain[response].asfactor() 
gbm = H2OGradientBoostingEstimator()

# Train
gbm.train(x=features, y=response, training_frame=dataTrain)
\end{lstlisting}
\vspace{-6ex}
\caption{Criteo training pipeline implemented in H20.}
\vspace{-3ex}
\label{fig:criteo-h20}
\end{figure}

\stitle{Amazon.}
In this pipeline first we featurize the text column of the dataset and then applies a linear classifier. 
For the featurization part, we used the \texttt{TextFeaturizer} transform in \tool and the \texttt{TfidfVectorizer} in \skl.
The \texttt{Text} \texttt{Featurizer/Tfidf} \texttt{Vectorizer} extracts numeric features from the input corps by producing a matrix of token ngrams counts.
As parameters, in both case we used word ngrams of size 1 and 2, and char ngrams of size 1, 2 and 3.
H2O implements the Skip-Gram word2vec model~\cite{word2vec} as the only text featurizer.
The training scripts are shown in Figure~\ref{fig:amazon-ml}, Figure~\ref{fig:amazon-sci}, and Figure~\ref{fig:amazon-h20}.

\begin{figure}[h]
\vspace{-2ex}
\lstset{numbersep=5pt, xleftmargin=10pt,   framexleftmargin=10pt}
\begin{lstlisting}
import pandas as pd
from nimbusml.linear_model import AveragedPerceptronBinaryClassifier
from nimbusml.feature_extraction.text import NGramFeaturizer
from nimbusml.feature_extraction.text.extractor import Ngram
from nimbusml import Pipeline

def LoadData(fileName):
    data = pd.read_csv(fileName, sep='\t', error_bad_lines=False)
    return data.iloc[:,1],data.iloc[:,0]  

# Load data
dataTrain, labelTrain = LoadData(train_file)

# Create pipeline
nimbus_ppl = Pipeline([
    NGramFeaturizer(word_feature_extractor = Ngram(ngram_length = 2), char_feature_extractor = Ngram(ngram_length = 3)),
    AveragedPerceptronBinaryClassifier()])
                 
# Train   
nimbus_ppl.fit(dataTrain,labelTrain,parallel = n_thread)
\end{lstlisting}
\vspace{-6ex}
\caption{Amazon pipeline implemented in NimbusML with Data  Frames.}
\vspace{-3ex}
\label{fig:amazon-ml}
\end{figure}

\begin{figure}[h]
\vspace{-2.5ex}
\lstset{numbersep=5pt, xleftmargin=10pt,   framexleftmargin=10pt}
\begin{lstlisting}
import pandas as pd
from sklearn.feature_extraction.text import TfidfVectorizer
from sklearn.pipeline import Pipeline, make_pipeline, FeatureUnion
from sklearn.linear_model import Perceptron

def LoadData(fileName):
    data = pd.read_csv(fileName, sep='\t', error_bad_lines=False)
    return data.iloc[:,1],data.iloc[:,0]  

# Load daa
dataTrain, labelTrain = LoadData(train_file)

# Create pipeline
wg = make_pipeline(TfidfVectorizer(analyzer='word', ngram_range = (1,2))) 
cg = make_pipeline(TfidfVectorizer(analyzer='char', ngram_range = (1,3)))

sk_ppl = Pipeline([  
   ('union', FeatureUnion(
             transformer_list=[('wordg',wg), ('charg',cg)])),
   ('ap', Perceptron())]) 
    
# Train
sk_ppl.fit(dataTrain,labelTrain)
\end{lstlisting}
\vspace{-6ex}
\caption{Amazon pipeline implemented in \skl.}
\label{fig:amazon-sci}
\end{figure}

\begin{figure}[!t]
\vspace{-2ex}
\lstset{numbersep=5pt, xleftmargin=10pt,   framexleftmargin=10pt}
\begin{lstlisting}
import h2o
from h2o.estimators.word2vec import Word2vecEstimator
from h2o.estimators.glm import GeneralizedLinearEstimator

h2o.init(nthreads=n_thread, min_mem_size='20g')

# Load data
dataTrain = h2o.import_file(train_file,header = 1, sep = "\t")

def tokenize(sentences):
    tokenized = sentences.tokenize("\\W+")
    return tokenized

# Create featurization pipeline
wordsTrain = tokenize(dataTrain["text"])
w2v_model = Word2vecEstimator(sent_sample_rate = 0.0, epochs = 5)

# Train featurizer
w2v_model.train(training_frame=wordsTrain)

# Create model pipeline
wordsTrain_vecs = w2v_model.transform(wordsTrain, aggregate_method = "AVERAGE")
dataTrain = dataTrain.cbind(wordsTrain_vecs)
glm_model = GeneralizedLinearEstimator(family= "binomial", lambda_ = 0, compute_p_values = True)

# Train
glm_model.train(wordsTrain_vecs.names, 'Label', dataTrain)
\end{lstlisting}
\vspace{-5ex}
\caption{Amazon pipeline implemented in H20.}
\label{fig:amazon-h20}
\vspace{-2ex}
\end{figure}

\stitle{Flight Delay.}
Here we pre-process all the feature columns into numeric values and we compare the performance of a single operator in ML.NET/NimbusML-DV/NimbusML-DF versus Sklearn/H2O without a pipeline (\texttt{LightGBM vs GradientBoosting}).
The NimbusML, \skl and H20 pipelines are reported in Figure~\ref{fig:fd-nm}, Figure~\ref{fig:fd-sci}, and Figure~\ref{fig:fd-h20}, respectively.

\stitle{Scale-up Experiments}
For this set of experiments we used the exact same scripts introduced above, except that we properly change the \texttt{n\_thread} parameter to set the number of used cores.

\begin{figure}[h]
\vspace{-2ex}
\lstset{numbersep=5pt, xleftmargin=10pt,   framexleftmargin=10pt}
\begin{lstlisting}
import pandas as pd
from nimbusml.ensemble import LightGbmRegressor

# Load data
df_train = pd.read_csv(train_file, header = None)
df_train.columns = df_train.columns.astype(str)

# Create pipeline
nimbus_m = LightGbmRegressor()

# Train
nimbus_m.fit(df_train.iloc[:,1:], df_train.iloc[:,0],parallel = n_thread)
\end{lstlisting}
\vspace{-5ex}
\caption{Flight Delay pipeline implemented in NimbusML using the Data Frame API.}
\label{fig:fd-nm}
\vspace{-2ex}
\end{figure}

\begin{figure}[h]
\vspace{-2ex}
\lstset{numbersep=5pt, xleftmargin=10pt,   framexleftmargin=10pt}
\begin{lstlisting}
import pandas as pd
from sklearn.ensemble import GradientBoostingRegressor
from sklearn.metrics import mean_squared_error

# Load data
df_train = pd.read_csv(train_file, header = None)
df_train.columns = df_train.columns.astype(str)

# Create pipeline
sklearn_m = GradientBoostingRegressor()

# Train
sklearn_m.fit(df_train.iloc[:,1:], df_train.iloc[:,0])
\end{lstlisting}
\vspace{-5ex}
\caption{Flight Delay pipeline in \skl.}
\label{fig:fd-sci}
\vspace{-4ex}
\end{figure}

\begin{figure}[b]
\vspace{-2ex}
\lstset{numbersep=5pt, xleftmargin=10pt,   framexleftmargin=10pt}
\begin{lstlisting}
import h2o
from h2o.estimators.gbm import H2OGradientBoostingEstimator

h2o.init(nthreads=n_thread,min_mem_size='20g')

# Load data
dataTrain = h2o.import_file(train_file,header = 0)

# Create pipeline
response = "C1"
features = ["C" + str(x) for x in range(2,41)]
gbm = H2OGradientBoostingEstimator()

# Train
gbm.train(x=features, y=response, training_frame=dataTrain)
\end{lstlisting}
\vspace{-5ex}
\caption{Flight Delay pipeline implemented in H20.}
\label{fig:fd-h20}
\vspace{-2ex}
\end{figure}